\documentclass[10pt,journal,compsoc]{IEEEtran}
\ifCLASSOPTIONcompsoc
  \usepackage[nocompress]{cite}
\else
  \usepackage{cite}
\fi
\ifCLASSINFOpdf
\else
\fi

\usepackage{hyperref}       
\usepackage{url}            
\usepackage{booktabs}       
\usepackage{nicefrac}       
\usepackage{microtype}      
\usepackage{subfigure}
\usepackage{makecell}

\usepackage{enumitem}
\usepackage{amsmath}
\usepackage{enumerate}
\usepackage{xcolor}
\usepackage{bm}
\usepackage{graphicx}
\usepackage{color}
\usepackage{chngpage}
\usepackage{import}
\usepackage{amssymb,amsfonts}
\usepackage{algorithmic}
\usepackage{textcomp}
\usepackage{flushend}
\usepackage{hhline}
\usepackage{multirow}
\usepackage{array}
\usepackage{gensymb}
\usepackage{nameref}
\usepackage{tabularx}
\usepackage[normalem]{ulem}
\graphicspath{{E:/papers/TPAMI/figs/}}
\DeclareRobustCommand\nobreakspace{\leavevmode\nobreak\ }
\newcolumntype{"}{!{\vrule width 1pt}}
\newcolumntype{L}[1]{>{\raggedright\let\newline\\\arraybackslash\hspace{0pt}}m{#1}}
\newcolumntype{C}[1]{>{\centering\let\newline\\\arraybackslash\hspace{0pt}}m{#1}}
\newcolumntype{R}[1]{>{\raggedleft\let\newline\\\arraybackslash\hspace{0pt}}m{#1}}

\def\allinone{all_in_one}
\begin{document}
\title{GaitSet: Cross-view Gait Recognition through Utilizing Gait as a Deep Set}

\author{\normalsize{
Hanqing Chao,
Kun Wang, 
Yiwei He,
Junping Zhang,~\IEEEmembership{Member,~IEEE},
Jianfeng Feng
\IEEEcompsocitemizethanks{\IEEEcompsocthanksitem This work was supported in part by Shanghai Municipal Science and Technology Major Project (Grant No. 2018SHZDZX01) and ZJLab, the National Key R \& D Program of China (No. 2018YFB1305104), and National Natural Science Foundation of China (Grant No. 61673118).}
\IEEEcompsocitemizethanks{\IEEEcompsocthanksitem Manuscript received xxx xx, 2020; revised xxx xx, 2020.}
\IEEEcompsocitemizethanks{\IEEEcompsocthanksitem Hanqing Chao, Kung Wang, Yiwei He, and Junping Zhang are with the Shanghai Key Laboratory of Intelligent Information Processing, School of Computer Science, Fudan University, Shanghai, 200438, P.R. China; Tel.: +86-21-55664503, Fax: +86-21-65654253, Email: \{hqchao16, KunWang17, heyw15, jpzhang\}@fudan.edu.cn}
\IEEEcompsocitemizethanks{\IEEEcompsocthanksitem Jianfeng Feng is with the Institute of Science and Technology for Brain-inspired Intelligence, Fudan University, Email: jianfeng64@gmail.com}
}
\IEEEcompsocitemizethanks{\IEEEcompsocthanksitem Corresponding author: Junping Zhang}
}

\markboth{IEEE TRANSACTIONS ON Pattern Analysis and Machine Intelligence, VOL. XX, NO. XX, XXX 2020}%
{Shell \MakeLowercase{\textit{et al.}}: GaitSet: Cross-view Gait Recognition through Utilizing Gait as a Deep Set}

\IEEEtitleabstractindextext{%
\begin{abstract}
Gait is a unique biometric feature that can be recognized at a distance; thus, it has broad applications in crime prevention, forensic identification, and social security. To portray a gait, existing gait recognition methods utilize either a gait template which makes it difficult to preserve temporal information, or a gait sequence that maintains unnecessary sequential constraints and thus loses the flexibility of gait recognition. In this paper, we present a novel perspective that utilizes gait as a \textbf{deep set}, which means that a set of gait frames are integrated by a global-local fused deep network inspired by the way our left- and right-hemisphere processes information to learn information that can be used in identification. Based on this \textbf{deep set} perspective, our method is \textbf{immune to frame permutations}, and can naturally \textbf{integrate frames from different videos} that have been acquired under different scenarios, such as diverse viewing angles, different clothes, or different item-carrying conditions. Experiments show that under normal walking conditions, our single-model method achieves an average rank-1 accuracy of 96.1\% on the CASIA-B gait dataset and an accuracy of 87.9\% on the OU-MVLP gait dataset. Under various complex scenarios, our model also exhibits a high level of robustness. It achieves accuracies of 90.8\% and 70.3\% on CASIA-B under bag-carrying and coat-wearing walking conditions respectively, significantly outperforming the best existing methods. Moreover, the proposed method maintains a satisfactory accuracy even when only small numbers of frames are available in the test samples; for example, it achieves 85.0\% on CASIA-B even when using only 7 frames. The source code has been released at \url{https://github.com/AbnerHqC/GaitSet}.
\end{abstract}

}

\maketitle

\IEEEdisplaynontitleabstractindextext

\IEEEpeerreviewmaketitle

\ifx\allinone\undefined
\input{../preamble}
\begin{document}
\fi

\IEEEraisesectionheading{\section{Introduction}\label{sec:introduction}}

\IEEEPARstart{U}{nlike} other biometric identification sources such as a face, fingerprint, or iris,  gait is a unique biometric feature that can be recognized from a distance without any intrusive interactions with subjects. This characteristic gives gait recognition high potential for use in applications such as crime prevention, forensic identification, and social security.

However, a person's variational poses in walking, which forms the basic information for gait recognition, is easily affected by exterior factors such as the subject's walking speed, clothing, and item-carrying condition as well as the camera's viewpoint and frame rate. These factors make gait recognition very challenging, especially cross-view gait recognition, which seeks to identify gait that might be captured from different angles. It thus is crucial to develop a practical gait recognition system.

The existing works have tried to tackle the problem from two aspects. 
They either regard gait as a single image or regard it as a video sequence. Methods in the first category compress all gait silhouettes into one image, \textit{i.e.}, a gait template, for gait recognition~\cite{liu2006improved,hu2013view,guan2014reducing,takemura2017input,chen2017multi,he2019multi,wu2017comprehensive}. 
Although various existing gait templates~\cite{chen2017multi,he2019multi,wu2017comprehensive} encode information as abundantly as possible, the compression process omits significant features such as temporal information and fine-grained spatial information. To address this issue, the methods in the second category extract features directly from the original gait silhouette sequences~\cite{liao2017pose,takemura2018multi,wu2019spatial}. These methods 
preserve more temporal information but would suffer a significant degradation when an input contains discontinuous frames or has a frame rate different from the training dataset.

\begin{figure}[t]
\centering
\includegraphics[width=1\linewidth, clip=true, trim=20 160 20 160]{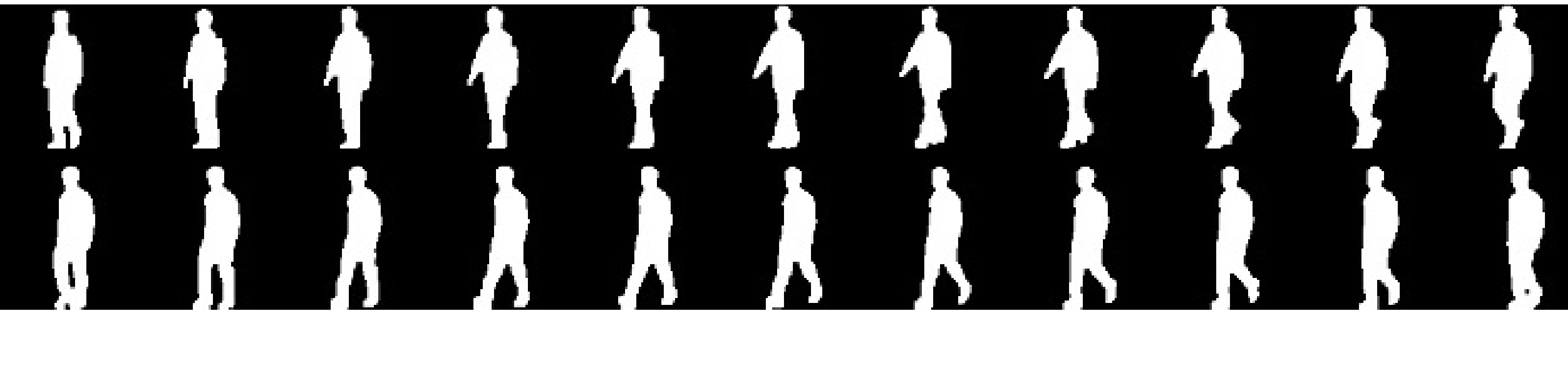}
\caption{From top-left to bottom-right are silhouettes of a
completed period of a subject in the \textbf{CASIA-B} gait dataset.}
\label{fig:gaitseq}
\end{figure}

To solve these problems, we present a novel perspective that regards gait as a set of gait silhouettes. Because gait is a periodic motion, it can be represented by a single period. Meanwhile, in a silhouette sequence containing one gait period, it can be observed that the silhouette in each position has a unique pose, as shown in Fig.~\ref{fig:gaitseq}. Given anyone's gait silhouettes, we can easily rearrange them into the correct order solely by observing their appearance. This suggests that the order of the poses in a gait period is not a key information to differentiate one person from others, since the pattern of the order is universal. Based on such an assumption, 
we can directly regard gait as a set of images and extract temporal information without ranking each frame like a video.

From this perspective, we propose an end-to-end deep learning model called Gaitset that extracts features from a gait frame set to identify gaits. Fig.~\ref{fig:pipeline} shows the overall scheme of Gaitset. The input to our model is a set of gait silhouettes. First, a CNN is used to extract frame-level features from each silhouette independently (local information). Second, an operation called Set Pooling is used to aggregate frame-level features into a single \textbf{set-level} feature (global information). Because this operation is conducted using  high-level feature maps instead of the original silhouettes, it preserves spatial and temporal information better than a gait template; this aspect is experimentally validated in Sec.~\ref{sec:ms}. The global-local fused deep network resembles the way our brain processes information~\cite{PBSM2004}.
Third, a structure called Horizontal pyramid mapping (HPM) is applied to project the set-level feature into a more discriminative space to obtain a final deep set representation.
The superiority of the proposed method can be summarized into the following three aspects:
\begin{itemize}
\item \textbf{Flexible:} Our model is pretty flexible since it imposes no constraints on the input except the size of the silhouette. This means that the input set can consist of any number of nonconsecutive silhouettes filmed under different viewpoints with different walking conditions. 
Related experiments are presented in Sec.~\ref{sec:pra}
\item \textbf{Fast:} Our model directly learns the deep set gait representation of gait instead of measuring the similarity between a pair of gait templates or sequences. Thus, the representation of each sample only needs to be computed once and the recognition can be done by comparing Euclidean distance between the representations of different samples.
\item \textbf{Effective:} Our model substantially improves the state-of-the-art performance on the CASIA-B~\cite{yu2006framework} and the OU-MVLP~\cite{Takemura2018} datasets, exhibiting strong robustness to view and walking condition variations and high generalization ability to large datasets.
\end{itemize}

Compared with our previous AAAI-19 conference paper on this topic, we have extended our work in four ways: 1) we surveyed and compared more state-of-the-art gait recognition algorithms; 2) we conducted more comprehensive experiments to evaluate the performance of the proposed GaitSet model; 3) we achieved better performance by improving the loss function used in GaitSet; 4) a post feature dimension reduction module were included to enhance the practicality.

\ifx\allinone\undefined
\newpage
\bibliographystyle{aaai}
\bibliography{../ref}
\end{document}
\fi
\ifx\allinone\undefined
\input{../preamble}
\begin{document}
\fi

\section{Related Works}
\label{sec:rw}
In this section, we briefly survey developments in gait recognition and set-based deep learning methods. 
\subsection{Gait Recognition}
Gait recognition can be broadly categorized into template-based and sequence-based approaches. In the former category, the previous works have generally divided this pipeline into two parts, i.e., template generation and matching. The goal of template generation is to compress gait information into a single image, e.g., a Gait Energy Image~(GEI)~~\cite{han2006individual} or a Chrono-Gait Image~(CGI)~\cite{wang2012human}. To generate a template, these approaches first estimate the human silhouettes in each frame through background removal. Then, they generate a gait template by applying pixel level operators to the aligned silhouettes~\cite{wang2012human}. In the template matching procedure, 
they first extract the gait representation from a template image using machine learning approaches such as canonical correlation analysis~(CCA)~\cite{xing2016complete}, linear discriminant analysis~(LDA)~\cite{liu2006improved,bashir2010gait} and deep learning~\cite{shiraga2016geinet}. Then, they measure the similarity between pairs of representations using Euclidean distance or other metric learning approaches~\cite{liu2006improved,guan2014reducing,wu2017comprehensive,takemura2017input}. For example, the view transformation model~(VTM) learns a projection between different views~\cite{makihara2006gait}; \cite{hu2013view} proposed view-invariant discriminative projection~(ViDP) to project the templates into a latent space to learn a view-invariant representation. Finally, they assign a label to the template based on the measured distance using a classifier, e.g., 
SVM or nearest neighbor classifier~\cite{yu2017gaitgan,he2019multi,takemura2017input,shiraga2016geinet,yu2017invariant,wu2017comprehensive}.

In the second category, the video-based approaches directly take a sequence of silhouettes as an input. For instance, the 3D CNN-based approaches [7], [9] extract temporal-spatial information using 3D convolutions; Liao [8]  and An [c] utilize human skeletons to lean gait features which is robust to the change of clothing; [23] fused sequential frames by LSTM attention units; and [10] proposed the spatial-temporal graph attention network (STGAN) to uncover the graph relationships between gait frames, followed by obtaining attention for gait video. In recent, a part-based model [a] was proposed to capture spatial-temporal feature of each part and achieved promising results. To prevent redundancy feature in part-based model, two stage training strategy was used in [b] to learn compact features successfully.

In the second category, the video-based approaches directly take a sequence of silhouettes as an input. For instance, the 3D CNN-based approaches~\cite{takemura2018multi,wu2017comprehensive} extract temporal-spatial information using 3D convolutions; Liao et al.~\cite{liao2017pose} and An et al.~\cite{an2020performance} utilized human skeletons to learn gait features robust to the change of clothing; Zhang et al.~\cite{zhang2019cross} fused sequential frames by LSTM attention units; and Wu et al.~\cite{wu2019spatial} proposed the spatial-temporal graph attention network (STGAN) to uncover the graph relationships between gait frames, followed by obtaining attention for gait video. Recently, a part-based model~\cite{fan2020gaitpart} was proposed to capture spatial-temporal features of each part and achieved promising results. To prevent redundancy feature in part-based model, a two-stage training strategy was used in~\cite{hou2020gait} to learn compact features effectively.

\begin{figure*}[h!t]
\centering
\includegraphics[width=1\linewidth, clip=true, trim=0 141 0 142]{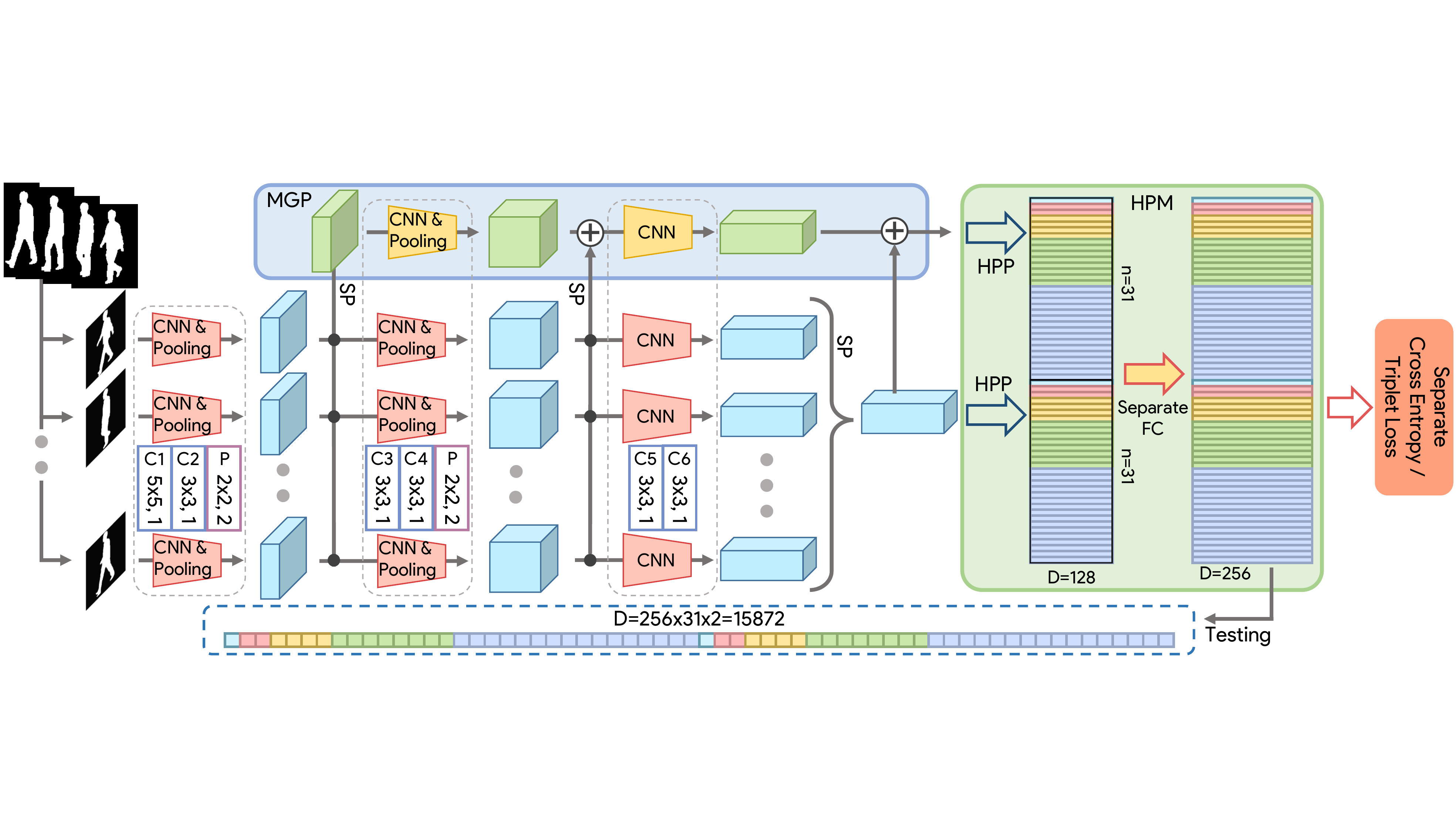}
\caption{The framework of GaitSet \cite{chao2019gaitset} .
`SP' represents set pooling. Trapezoids represent convolution and pooling blocks and those in the same column have the same configurations, as shown by the rectangles with capital letters. Note that although the blocks in MGP have the same configurations as those in the main pipeline, the parameters are shared only across blocks in the main pipeline -- not with those in MGP. HPP represents horizontal pyramid pooling~\protect\cite{fu2018horizontal}.}
\label{fig:pipeline}
\end{figure*}

\subsection{Deep Learning on an Unordered Set}
Most deep learning works have been focusing on regular input representations such as video sequences or images. The initial goal for using unordered sets was to address point cloud tasks in the computer vision domain~\cite{qi2017pointnet} based on PointNet. Using an unordered set, PointNet can avoid the noise introduced by quantization and the extension of data, leading to a high prediction performance. Since then, set-based methods have been widely used in the point cloud domain~\cite{wang2018dynamic,zhou2017voxelnet,qi2017pointnet++} and in content recommendation~\cite{hamilton2017inductive} and image captioning~\cite{krause2017hierarchical} by integrating features into the form of a set. \cite{zaheer2017deep} further formalized the deep learning tasks defined for sets and characterized 
of the permutations using invariant functions. To the best of our knowledge, this topic has not been studied in depth in the gait recognition domain except in our previous AAAI-19 conference version of this paper~\cite{chao2019gaitset}.

\ifx\allinone\undefined
\newpage
\bibliography{../ref}
\end{document}
\fi
\ifx\allinone\undefined
\input{../preamble}
\begin{document}
\fi

\section{GaitSet}
\label{sec:method}

In this section, we introduce the details of our GaitSet method, which learns deep discriminative information from a set of gait silhouettes. To improve understanding, the overall pipeline is illustrated in Fig.~\ref{fig:pipeline}.

\subsection{Problem Formulation}
The concept for regarding gait as a deep set will be formulated first. Given a dataset of $N$ people with identities $y_i, i\in{1,2, ..., N}$, we assume that the gait silhouettes of a certain person are subject to a distribution $\mathcal{P}_i$ which is uniquely related to that individual. Therefore, all silhouettes in one or more sequences of a given person can be regarded as a set of $n$ silhouettes $\mathcal{X}_i=\{x_i^j|j=1,2,...,n\}$, where $x_i^j\sim\mathcal{P}_i$. 
Under this assumption, we tackle the gait recognition task via 3 steps, formulated as follows:
\begin{equation}
f_i=H(G(F(x_i^1), F(x_i^2), ..., F(x_i^n)))
\end{equation}
where $F$ is a convolutional network that seeks to extract frame-level features from each gait silhouette. The function $G$ is a permutation invariant function used to map a set of frame-level features to a set-level feature~\cite{zaheer2017deep} based on set pooling~(SP), which will be introduced in Sec.~\ref{sec:SP}. The function $H$ learns the deep set discriminative representation of $\mathcal{P}_i$ from the set-level feature through a structure called horizontal pyramid mapping~(HMP) which will be discussed in Sec.~\ref{sec:HPM}. The input $\mathcal{X}_i$ is a tensor with four dimensions: set dimension, image channel dimension, image height dimension, and image width dimension.

\subsection{Set Pooling}\label{sec:SP}
The goal of Set Pooling~(SP) is to condense a set of gait information, formulated as $z=G(V)$, 
where $z$ denotes the set-level feature and  $V=\{v^j|j=1,2,...,n\}$ denotes the frame-level features, where $v^j$ means the $j$-th frame-level feature map and $n$ denotes the number of gait frames in a set. Note that there are two constraints when performing an SP operation. First, if we expect to take a set as an input, the function should be a permutation invariant function satisfying:
\begin{equation}
\label{eq:sp}
G(\{v^j|j=1,2,...,n\}) = G(\{v^{\pi(j)}|j=1,2,...,n\})
\end{equation}
where $\pi$ is any permutation~\cite{zaheer2017deep}. Second, the function $G$ should be able to take a \textbf{set} with arbitrary cardinality because the number of a person's gait silhouettes can be arbitrary in a real-world scenario. Next, we describe several instantiations of $G$. The experiments will show that although different instantiations of SP do influence the performances, they do not produce significant differences, and all SP instantiations achieve better performances than do the GEI-based methods by a large margin.

\noindent \textbf{Basic Statistical Functions}~~
To meet the invariant constraint requirement in Eq.~\ref{eq:sp}, one rational strategy of SP is to use statistical functions on the set dimension. To balance the representativeness and the computational cost, we considered three statistical functions: $\mathrm{max}(\cdot)$, $\mathrm{mean}(\cdot)$ and $\mathrm{median}(\cdot)$. A comprehensive comparison will be given in Sec.~\ref{sec:exp}.

\noindent \textbf{Joint Functions}~~
Further, two feasible ways to join the aforementioned $3$ basic statistical functions are analyzed as follows:
\begin{align}
\label{eq:sum}
G(\cdot)&=\mathrm{max}(\cdot)+\mathrm{mean}(\cdot)+\mathrm{median}(\cdot) \\
\label{eq:cat}
G(\cdot)&=\mathrm{1\_1C}(\mathrm{cat}(\mathrm{max}(\cdot),
 \mathrm{mean}(\cdot), \mathrm{median}(\cdot)))
\end{align}
where $\mathrm{cat}$ means concatenating on the channel dimension, $1\_1C$ denotes $1\times1$ convolutional layer, and $\mathrm{max}$, $\mathrm{mean}$ and $\mathrm{median}$ are applied to the set dimension. Eq.~\ref{eq:cat} is an extended version of Eq.~\ref{eq:sum}, which allows the $1\times1$ convolutional layer to learn a proper weight for combining information extracted by different statistical functions.

\begin{figure*}[htbp]
\centering
\includegraphics[width=1\linewidth, clip=true, trim=0 110 0 0]{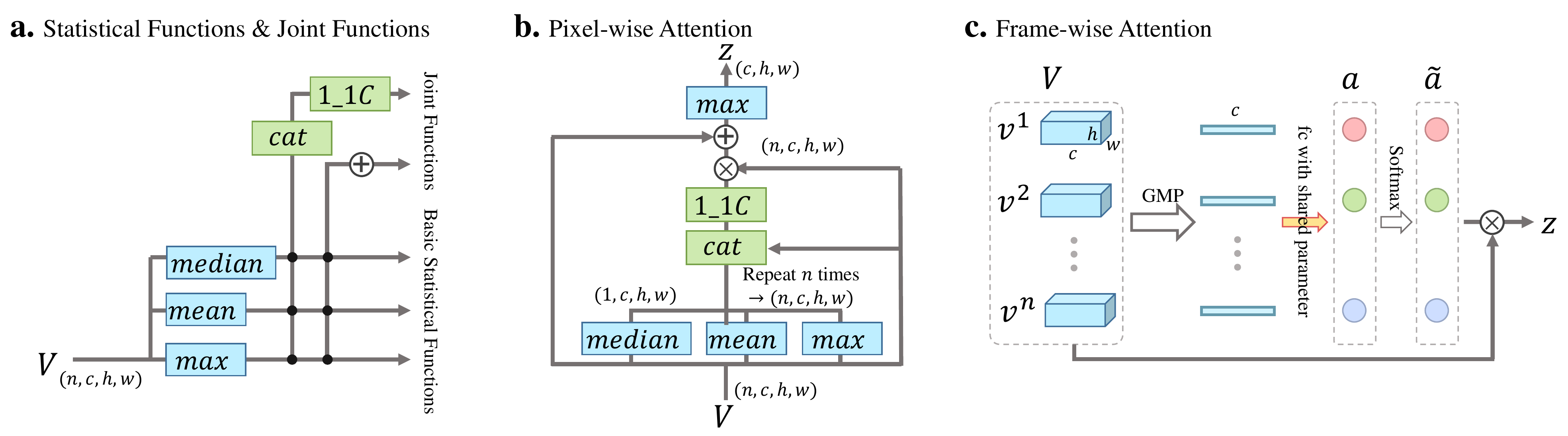}
\caption{Seven different instantiations of Set Pooling~(SP). $1\_1C$ and $cat$ represent the $1\times1$ convolutional layer and the concatenate operation, respectively. Here, $n$ represents the number of feature maps in a set, and $c$, $h$ and $w$ denote the number of channels, the height and width of a feature map, respectively. \textbf{a.} Three basic statistical SP and two joint SP. \textbf{b.} Pixel-wise attention SP. \textbf{c.} Frame-wise attention SP.}
\label{fig:attention}
\end{figure*}

\noindent \textbf{Attention}~~
Visual attention has been successfully applied in many computer vision tasks~\cite{wang2018non,xu2015show,li2018harmonious}, and we also capitalize on attention to implement SP. We included two attention strategies in our work. The first one is a \textbf{pixel-wise attention}. Specifically, we refine the output of SP by utilizing the global information to learn an element-wise attention map for each frame-level feature map, as shown in Fig.~\ref{fig:attention}b. 
First, global information is first collected by the statistical functions. Then, it is input into a $1\times 1$ convolutional layer along with the original feature map to calculate an attention map for the refinement. The final set-level feature $z$ is extracted by employing a MAX operation on the set of the refined frame-level feature maps. 
We use the residual structure to accelerate and stabilize the convergence. Another is a \textbf{frame-wise attention}, in which global max pooling is first applied on each $v^j$ to get a compressed frame-wise feature. Then, based on the frame-wise feature, a fully connected layer is applied to calculate a frame-wise weight $a^j$. Finally, $z$ is calculated by $\sum_{j=1}^n \tilde{a}^j v^j$, where $\tilde{a}^j$ is the softmax-normalized frame-wise weight. Fig.~\ref{fig:attention}c illustrates the architecture of the frame-wise attention.

\subsection{Horizontal Pyramid Mapping}\label{sec:HPM}
\begin{figure}[htbp]
\centering
\includegraphics[width=1\linewidth, clip=true, trim=70 80 70 100]{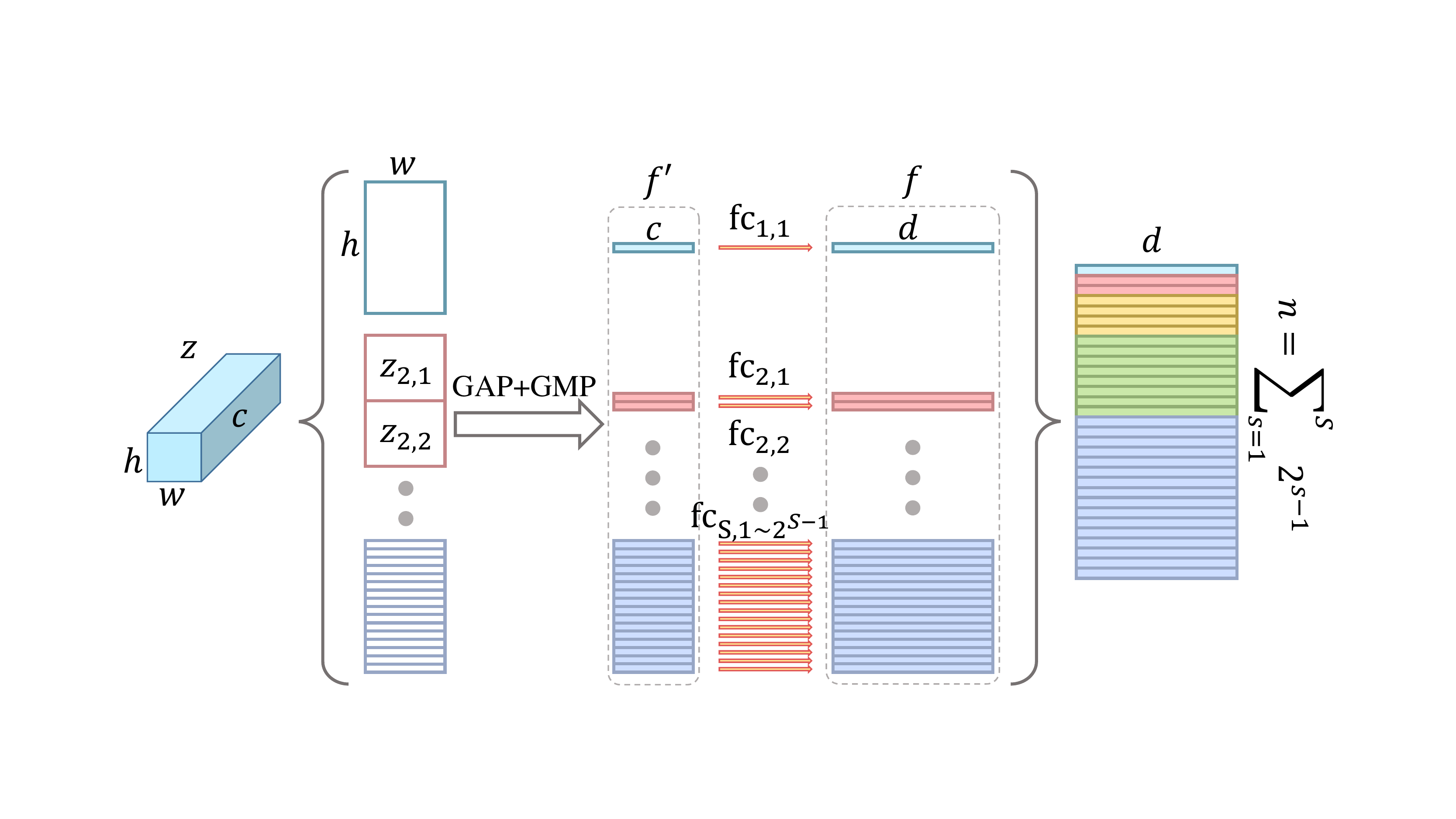}
\caption{The structure of horizontal pyramid mapping \cite{chao2019gaitset}.}
\label{fig:hpm}
\end{figure}

In the literature, splitting feature maps into strips is a commonly used tactic in person re-identification  tasks~\cite{wang2018learning,fu2018horizontal}. For instance, \cite{fu2018horizontal} proposed horizontal pyramid pooling (HPP) through cropping and resizing the images into a uniform size based on pedestrian size while varying the discriminative parts from image to image. With 4 scales, HPP can thus help the deep network gather both local and global information by focusing on features with different sizes. Here, we improve HPP to adapt it to the gait recognition task; instead of applying a $1\times1$ convolutional layer after the pooling, we use independent fully connected layers~(FC) for each pooled feature to map it into the discriminative space, as shown in Fig.~\ref{fig:hpm}. We call this approach horizontal pyramid mapping (HPM).

Concretely, HPM has $S$ scales. On scale $s\in{1,2,...,S}$, the feature map extracted by SP is split into $2^{s-1}$ strips on the height dimension, i.e. $\sum_{s=1}^S2^{s-1}$ strips in total. Then, global pooling is applied to the 3-D strips to obtain 1-D features. For a strip $z_{s,t}$ where $t\in1,2,...,2^{s-1}$ represents the index of the strip in the scale, the global pooling is formulated as 
$f'_{s,t}=maxpool(z_{s,t})+avgpool(z_{s,t})$, 
where $maxpool$ and $avgpool$ denote global max pooling (GMP) and global average pooling (GAP) respectively. Note that the functions $maxpool$ and $avgpool$ are used at the same time because the combined results outperform applying either operation alone. The final step is to employ FCs to map the features $f'$ into a deep discriminative space. Because the strips at different scales depict features of different receptive fields, and different strips at each scale depict features of different spatial positions, using independent FCs is a natural choice, as shown in Fig.~\ref{fig:hpm}.

\subsection{Multilayer Global Pipeline}\label{sec:MGP}

Generally, different layers of a convolutional network have different receptive fields. The deeper a layer is, the larger the receptive field will be. Thus, pixels in the feature maps of a shallow layer pay more attention to local and fine-grained information while those in deeper layers focus more on global and coarse-grained information. The set-level features extracted by applying SP to different layers are analogous. As shown in the main pipeline of Fig.~\ref{fig:pipeline}, only one SP is applied to the last layer of the convolutional network. To collect different-level \textbf{set} information, we propose a multilayer global pipeline~(MGP), which has a similar structure to the convolutional network in the main pipeline; however, we add the set-level features extracted by different layers to MGP. The final feature map generated by MGP is also mapped into $\sum_{s=1}^S2^{s-1}$ features by HPM. Note that 1) the HPM that executes after MGP does not share parameters with the HPM that executes after the main pipeline.
2) The main pipeline is similar to that of human cognition, which focuses intuitively on a person's profile, whereas the MGH can preserve more details of a person's walking movements. 

\subsection{Loss Functions and Training Strategy}
\label{sec:Loss}
In the field of identification~\cite{taigman2014deepface,schroff2015facenet,zheng2016person}, two loss functions are widely used, \textit{i.e.}, cross entropy loss and triplet loss~\cite{schroff2015facenet}. To obtain the best performance, we conducted comprehensive experiments on these two loss functions.

\noindent \textbf{Cross-entropy loss} is common in classification tasks. It measures the gap between a predictive distribution and the corresponding true distribution. In the recognition task, the output classes reflect all labels (identities) in the training set. As mentioned above, the output of the network are $2\times\sum_{s=1}^S2^{s-1}$ features with a dimension of $d$. During training, a cross entropy loss was calculated for each feature and then all losses were summed up as a total loss. During testing, the feature before the softmax layer is used for recognition. 

\noindent \textbf{Triplet loss} was initially proposed for face recognition~\cite{schroff2015facenet} but has become a popular loss function for metric embedding learning and has achieved high performances on various tasks~\cite{schroff2015facenet,hermans2017defense,chung2018voxceleb2,dong2018triplet,chao2019gaitset}. 
It aims to pull semantically-similar points close to each other while pushing semantically-different points away from each other~\cite{hermans2017defense}. The specific version of triplet loss adopted in this paper is Batch All~($BA_+$) triplet loss~\cite{hermans2017defense}. 

Specifically, we denote a sample triplet as $r=(\alpha, \beta, \gamma)$ where $\alpha$ represents the anchor, $\beta$ is a sample with the same label as the anchor $\alpha$, and $\gamma$ denotes a sample with a label different from that of the anchor $\alpha$. Then, the $BA_+$ triplet loss of this triplet is defined as follows:
\begin{align}
\label{eq:loss}
L(r) = ReLU(\xi+D_{\alpha, \beta}-D_{\alpha, \gamma}),
\end{align}
where $\xi$ is the margin between intraclass distance $D_{\alpha, \beta}$ and interclass distance $D_{\alpha, \gamma}$. For a triplet $r$, each sample has $2\times\sum_{s=1}^S2^{s-1}$ features. We calculated the triplet loss for each  corresponding feature triplet, \textit{i.e.}, $2\times\sum_{s=1}^S2^{s-1}$ triplet losses were calculated.

\noindent \textbf{Using a combination of two loss functions}
Our previous work~\cite{chao2019gaitset} only used Batch All~($BA_+$) triplet loss and achieved state-of-the-art performance. In this study, to improve the learning ability, we combined the cross entropy loss with the triplet loss. First, cross entropy loss was used to train the network to converge. Then, a smaller learning rate with Batch All~($BA_+$) triplet loss was used to let the model find a more discriminant metric space. Experiments to compare these two losses are shown in Sec.~\ref{loss strategies}. 

\subsection{Training and Test}\label{sec:TT}
\noindent \textbf{Training.}
In the training phase,
a batch with a size of $p\times k$ is sampled from the training set, where $p$ denotes the number of persons and $k$ is the number of training samples each person has in the batch.
Note that although the experiment shows that our model performs well when its input is the set composed of silhouettes gathered from arbitrary sequences, 
during the training phase, a sample is composed only of silhouettes sampled in one sequence.

\noindent \textbf{Testing.}
Given a query $\mathcal{Q}$, the goal is to retrieve all the \textbf{sets} with the same identity in gallery set $\mathbb{G}$. We denote a sample in $\mathbb{G}$ as $\mathcal{G}$. First, $\mathcal{Q}$ is input into GaitSet model to generate $2\times\sum_{s=1}^S2^{s-1}$ multiscale features, then all these features are concatenated into a final representation $\mathcal{F}_\mathcal{Q}$ as shown in Fig.~\ref{fig:pipeline}. The same process is applied to each $\mathcal{G}$ to obtain $\mathcal{F}_\mathcal{G}$. Finally, $\mathcal{F}_\mathcal{Q}$ is compared with every $\mathcal{F}_\mathcal{G}$ to calculate the rank-$1$ recognition accuracy, which means the percentage of the correct subjects ranked first, based on nearest Euclidean distance.

\subsection{Post Feature Dimension Reduction}\label{sec:FRed}
As introduced in Sec.~\ref{sec:TT}, the identification is achieved by comparing $\mathcal{F}_\mathcal{Q}$ with every $\mathcal{F}_\mathcal{G}$ to find the nearest neighbor. Let $d_f$ denotes the dimension of the final representation $\mathcal{F}$. The computational complexity of this identification process is $O(d_f|\mathbb{G}|)$ where $|\cdot|$ calculates the cardinality of a set. In practical applications, $|\mathbb{G}|$ could be extremely large. A small $d_f$ will be the key to keep the process efficient. Thus, we proposed a post feature dimension reduction module which is a post trained linear projection to reduce the dimension of the output feature while maintaining a competitive recognition accuracy. Specifically, based on a trained GaitSet model with fixed parameters, we feed the learned $d_f$ dimensional feature into a fully connected layer with an output of dimension $d_f'$. The new $d_f'$ dimensional features are used to calculate a triplet loss to train the fully connected layer. In Sec.~\ref{sec:fdr} we show that compared with directly shrinking the output dimension of the HPM, our post feature dimension reduction model can effectively reduce the feature dimension into a significantly smaller size while preserving a high recognition performance.

\ifx\allinone\undefined
\newpage
\bibliographystyle{aaai}
\bibliography{../ref}
\end{document}
\fi
\ifx\allinone\undefined
\input{preamble}
\begin{document}
\fi

\section{Experiments}\label{sec:exp}

In this section, we report the results of comprehensive experiments conducted to evaluate the performance of the proposed GaitSet. First, we compared GaitSet with other state-of-the-art methods on two public gait datasets: CASIA-B~\cite{yu2006framework} and OU-MVLP~\cite{Takemura2018}. Then, we conducted a set of ablation studies on CASIA-B.
Third, we studied the effectiveness of feature dimension reduction. Finally, we analyzed the practicality of GaitSet from three aspects: limited silhouettes, multiple views, and multiple walking conditions.

\subsection{Datasets}
The \textbf{CASIA-B}~dataset~\cite{yu2006framework} is a popular gait dataset that 
contains 124 subjects labeled from 001 to 124. Each subject has 3 walking conditions, \textit{i.e.}, normal~(NM) (6 video sequences per subject), walking with a bag~(BG) (2 video sequences per subject) and wearing a coat or jacket~(CL) (2 video sequences per subject). Each sequence is simultaneously framed under 11 views ($0^\circ, 18^\circ, ..., 180^\circ$). Thus, this dataset contains $124\times(6+2+2)\times11=113,640$ videos in total.
Since this dataset does not include official training and test subset partitions, we conducted our experiments using three kinds of division popular 
in the current literature. Based on the sizes of the training sets, we name these three kinds of division small-sample training~(ST), medium-sample training~(MT) and large-sample training~(LT). In ST, the first 24 subjects 
(001-024) were used for training and the remaining 100 subjects were used for testing with no overlap. In MT, the first 62 subjects (001-062) were used for training, and the remaining 62 subjects were used for testing. In LT, the first 74 subjects (001-074) were used for training and the remaining 50 subjects were used for testing. For the test sets in all three settings, the first 4 sequences of the NM condition~(i.e., NM \#1-4) were kept in the gallery, and the remaining 6 sequences were divided into 3 probe subsets, i.e. the NM subset condition \#5-6, the BG subset containing BG \#1-2 and the CL subset containing CL \#1-2.

\noindent The \textbf{OU-MVLP}~dataset~\cite{Takemura2018} is currently the largest public gait dataset. It contains 10,307 subjects with 14 views~($0^\circ, 15^\circ, ...,90^\circ;180^\circ,195^\circ,...,270^\circ$) per subject and 2 sequences~(\#00-01) per view. According to its protocol, 
the sequences of 5,154 subjects are used for training, and the sequences of the remaining 5,154 subjects are used for testing.
In the test set, sequences with index \#01 are kept in the gallery and those with index \#00 are used as probes.

\subsection{Parameter Setting}
In all the experiments, the input was a set of aligned silhouettes of size $64\times 44$. The silhouettes were directly provided by the datasets and were aligned based on the methods described in~\cite{Takemura2018}. We adopted the Adam optimizer~\cite{kingma2014adam} for training our GaitSet network.
The code for all the experiments was written in Python with Pytorch~0.4.0. The models were trained on a computer equipped with 4 NVIDIA 1080TI GPUs.
Unless otherwise stated, the set cardinality during the training phase was set to $30$. 
The margin $\xi$ in the triplet loss function (Eq.~\ref{eq:loss}) was set to $0.2$. The number of HPM scales $S$ was set to $5$.
\textbf{1)}, For the \textbf{CASIA-B} dataset, we set the number of channels in $C1$ and $C2$ as 32, in $C3$ and $C4$ as 64, and in $C5$ and $C6$ as 128. Under these settings, the average computational complexity of our model is 8.6GFLOPs. \textbf{2)}~On the \textbf{OU-MVLP} dataset, which contains 20 times more sequences than CASIA-B, we used convolutional layers with more channels, i.e., $C1=C2=64, C3=C4=128, C5=C6=256$.
For convenience, the details of batch size, learning rate, and training iterations on different experimental settings are listed in Tab.~\ref{tab:train_detail}. Furthermore, rank-1 accuracy is adopted as a criterion in the subsequent evaluations.

\begin{table*}[htbp]
  \centering
  \caption{Batch size $(B_S)$, learning rate $(L_R)$, and training iterations (Iter) on OU-MVLP and the three settings of CASIA-B (CASIA-ST, CASIA-MT, and CASIA-LT). When being trained with cross entropy loss, a mini-batch is randomly selected from the training set. When being trained with triplet loss, a mini-batch is composed as described in Sec.~\ref{sec:TT}. 
  }
   \begin{tabular}{|p{0.8cm}|c|c|c|c|c|}
            \hline
      && CASIA-ST & CASIA-MT & CASIA-LT & OU-MVLP \\
      \hline
      \multirow{3}{*}{CE} & $B_S$ & 128  & 128  & 128 & 512 \\
      \cline{2-6}
      & $L_R$ & 1e-4  & 1e-4  & 1e-4 & Iter$<$150k: 1e-4; Iter$>=$150k: 1e-5 \\
      \cline{2-6}
      & Iter & 70k & 80k & 90k & 800k \\
      \hline
      \multirow{3}{*}{\shortstack{Triplet \\ Only}}& $B_S$ & $p=8$, $k=16$  & $p=8$, $k=16$  & $p=8$, $k=16$ & $p=32$, $k=16$ \\
      \cline{2-6}
      & $L_R$ & 1e-4  & 1e-4  & 1e-4 & Iter$<$150k: 1e-4; Iter$>=$150k: 1e-5 \\
      \cline{2-6}
      & Iter & 80k & 80k & 100k & 800k \\
      \hline
      \multirow{3}{*}{\shortstack{Triplet \\ Tune}}& $B_S$ & $p=8$, $k=16$  & $p=8$, $k=16$  & $p=8$, $k=16$ & $p=32$, $k=16$ \\
      \cline{2-6}
      & $L_R$ & 1e-5  & 1e-5  & 1e-5 & Iter$<$600k: 1e-4; Iter$>=$600k: 1e-5 \\
      \cline{2-6}
      & Iter & 30k & 40k & 60k & 700k \\
      \hline
      \end{tabular}
  \label{tab:train_detail}
\end{table*}

\subsection{Brief Introduction of Compared Methods}
Among our compared methods, View-invariant Discriminative Projection (ViDP)~\cite{hu2013view} uses a unitary
linear projection to project the templates intoa latent space to learn a view-invariant represent. Correlated Motion Co-Clustering (CMCC)~\cite{kusakunniran2014recognizing} first uses motion co-clustering to partition the most related parts of gaits from different views into the same group, and then applies canonical correlation analysis (CCA) on each group to maximize the correlation between gait information across views. Wu et al. proposed several CNN based models in~\cite{wu2017comprehensive}. CNN-LB feeds GEIs of two gait sequences into a 3-layer CNN as two channels and judges whether the two GEIs belong to the same person; CNN-3D runs 3-layers 3D-CNN on 9 adjacent frames and averages predictions of 16 9-frame samples to get the final output; CNN-Ensemble aggregates outputs of 8 different networks and achieves the best performance in this work. Yu et al.~\cite{yu2017invariant} applied AutoEncoder (AE) to extract view-invariant features. He et al.~\cite{yu2017invariant} proposed a multi-task GAN (MGAN) to project gait features from one angle to another angle for multi-view gait recognition. Angle Center Loss (ACL)~\cite{zhang2019cross} which is robust to different local parts and temporal window sizes is proposed to learn discriminative gait features. GEINet~\cite{shiraga2016geinet} classifies GEIs of different persons with a 2-layer CNN followed by 2 fully connected layers. Takemura et al.~\cite{takemura2017input} improved the structures proposed in~\cite{wu2017comprehensive} to leverage triplet loss.

\subsection{Main Results}
\label{sec:mr}

\subsubsection{CASIA-B}

\begin{table*}[htbp]
\centering
\caption{Averaged rank-1 accuracies on \textbf{CASIA-B} under three different experimental settings, excluding identical-view cases.}
\begin{tabularx}{\textwidth}{@{\extracolsep{\fill}}|c|c||c|p{0.4cm}|p{0.4cm}|p{0.4cm}|p{0.4cm}|p{0.4cm}|p{0.4cm}|p{0.4cm}|p{0.4cm}|p{0.4cm}|p{0.4cm}|p{0.4cm}|c|}
    \hline
    \multicolumn{3}{|c|}{Gallery NM\#1-4}&\multicolumn{11}{c|}{0\degree-180\degree}&\multirow{2}{*}{mean}\\ \cline{1-14}
    \multicolumn{1}{|c}{}&\multicolumn{1}{c}{Probe}&&0\degree&18\degree&36\degree&54\degree&72\degree&90\degree&108\degree&126\degree&144\degree&162\degree&180\degree&\\
    \hline \hline

    \multirow{8}{*}{\shortstack{ST \\ (24)}}&\multirow{6}{*}{NM\#5-6}&ViDP~\protect\cite{hu2013view} & $-$ & $-$ & $-$ & 59.1  & $-$ & 50.2  & $-$ & 57.5  & $-$ & $-$ & $-$ & $-$ \\
    &&CMCC~\protect\cite{kusakunniran2014recognizing} & 46.3  & $-$ & $-$ & 52.4  & $-$ & 48.3  & $-$ & 56.9  & $-$ & $-$ & $-$ & $-$ \\
    &&CNN-LB~\protect\cite{wu2017comprehensive}& 54.8  & $-$ & $-$ & 77.8  & $-$ & 64.9  & $-$ & 76.1  & $-$ & $-$ & $-$ & $-$ \\
    &&GaitSet(ours)&\textbf{71.6 } & \textbf{87.7 } & \textbf{92.6 } & \textbf{89.1 } & \textbf{82.4 } & \textbf{80.3 } & \textbf{84.4 } & \textbf{89.0 } & \textbf{89.8 } & \textbf{82.9 } & \textbf{66.6 } & \textbf{83.3 } \\
    \cline{2-15}
    &\multirow{1}{*}{BG\#1-2}&GaitSet(ours) & 64.1  & 76.4  & 81.4  & 82.4  & 77.2  & 71.8  & 75.4  & 80.8  & 81.2  & 75.7  & 59.4  & 75.1  \\
    \cline{2-15}
    &\multirow{1}{*}{CL\#1-2}&GaitSet(ours) & 36.4  & 49.7  & 54.6  & 49.7  & 48.7 & 45.2  & 45.5  & 48.2  & 47.2  & 41.4  & 30.6  & 45.2  \\
    \hline \hline

    \multirow{12}{*}{\shortstack{MT \\ (62)}}&\multirow{4}{*}{NM\#5-6}
    &AE~\protect\cite{yu2017invariant} & 49.3  & 61.5  & 64.4  & 63.6  & 63.7  & 58.1  & 59.9  & 66.5  & 64.8  & 56.9  & 44.0  & 59.3  \\
    &&MGAN~\protect\cite{he2019multi} & 54.9  & 65.9  & 72.1  & 74.8  & 71.1  & 65.7  & 70.0  & 75.6  & 76.2  & 68.6  & 53.8  & 68.1  \\
    &&GaitSet(ours) & \textbf{89.7 } & \textbf{97.9 } & \textbf{98.3 } & \textbf{97.4 } & \textbf{92.5 } & \textbf{90.4} & \textbf{93.4 } & \textbf{97.0 } & \textbf{98.9 } & \textbf{95.9 } & \textbf{86.6 } & \textbf{94.3 } \\
    \cline{2-15}

    &\multirow{4}{*}{BG\#1-2}
    &AE~\protect\cite{yu2017invariant} & 29.8  & 37.7  & 39.2  & 40.5  & 43.8  & 37.5  & 43.0  & 42.7  & 36.3  & 30.6  & 28.5  & 37.2  \\
    &&MGAN~\protect\cite{he2019multi} & 48.5  & 58.5  & 59.7  & 58.0  & 53.7  & 49.8  & 54.0  & 61.3  & 59.5  & 55.9  & 43.1  & 54.7  \\
    &&GaitSet(ours) & \textbf{79.9 } & \textbf{89.8 } & \textbf{91.2 } & \textbf{86.7 } & \textbf{81.6 } & \textbf{76.7 } & \textbf{81.0 } & \textbf{88.2 } & \textbf{90.3 } & \textbf{88.5 } & \textbf{73.0 } & \textbf{84.3 } \\
    \cline{2-15}
    &\multirow{4}{*}{CL\#1-2}
    &AE~\protect\cite{yu2017invariant} & 18.7  & 21.0  & 25.0  & 25.1  & 25.0  & 26.3  & 28.7  & 30.0  & 23.6  & 23.4  & 19.0  & 24.2  \\
    &&MGAN~\protect\cite{he2019multi} & 23.1  & 34.5  & 36.3  & 33.3  & 32.9  & 32.7  & 34.2  & 37.6  & 33.7  & 26.7  & 21.0  & 31.5  \\
    &&GaitSet(ours) & \textbf{52.0 } & \textbf{66.0 } & \textbf{72.8 } & \textbf{69.3 } & \textbf{63.1 } & \textbf{61.2 } & \textbf{63.5 } & \textbf{66.5 } & \textbf{67.5 } & \textbf{60.0 } & \textbf{45.9 } & \textbf{62.5 } \\
    \hline \hline

    \multirow{7}{*}{\shortstack{LT \\ (74)}}&\multirow{3}{*}{NM\#5-6}&CNN-3D~\protect\cite{wu2017comprehensive}&87.1  & 93.2  & 97.0  & 94.6  & 90.2  & 88.3  & 91.1  & 93.8  & 96.5  & 96.0  & 85.7  & 92.1  \\
    &&CNN-Ensemble~\protect\cite{wu2017comprehensive} &88.7  & 95.1  & 98.2  & 96.4  & 94.1  & 91.5  & 93.9  & 97.5  & 98.4  & 95.8  & 85.6  & 94.1  \\
    &&ACL~\protect\cite{zhang2019cross}  & \textbf{92.0} & 98.5 & \textbf{100.0} & \textbf{98.9} & \textbf{95.7}& 91.5 & 94.5 & 97.7 & 98.4 & 96.7 & \textbf{91.9} & 96.0 \\
    &&GaitSet(ours) & 91.1  & \textbf{99.0 } & 99.9  & 97.8  & 95.1  & \textbf{94.5 } & \textbf{96.1 } & \textbf{98.3 } & \textbf{99.2 } & \textbf{98.1 } & 88.0  & \textbf{96.1 } \\
    \cline{2-15}
    &\multirow{2}{*}{BG\#1-2}&CNN-LB~\protect\cite{wu2017comprehensive}&64.2  & 80.6  & 82.7  & 76.9  & 64.8  & 63.1  & 68.0  & 76.9  & 82.2  & 75.4  & 61.3  & 72.4  \\
    &&GaitSet(ours) & \textbf{86.7 } & \textbf{94.2 } & \textbf{95.7 } & \textbf{93.4 } & \textbf{88.9 } & \textbf{85.5 } & \textbf{89.0 } & \textbf{91.7 } & \textbf{94.5} & \textbf{95.9 } & \textbf{83.3 } & \textbf{90.8 } \\
    \cline{2-15}
    &\multirow{2}{*}{CL\#1-2}&CNN-LB~\protect\cite{wu2017comprehensive}&37.7  & 57.2  & 66.6  & 61.1  & 55.2  & 54.6  & 55.2  & 59.1  & 58.9  & 48.8  & 39.4  & 54.0  \\
    &&GaitSet(ours) & \textbf{59.5 } & \textbf{75.0 } & \textbf{78.3 } & \textbf{74.6 } & \textbf{71.4 } & \textbf{71.3 } & \textbf{70.8 } & \textbf{74.1 } & \textbf{74.6 } & \textbf{69.4 } & \textbf{54.1 } & \textbf{70.3 }  \\
    \hline

\end{tabularx}
\label{tab:casia-b}
\end{table*}

Tab.~\ref{tab:casia-b} shows a comparison between the state-of-the-art methods~\footnote{Since Wu et al proposed more than one model~\cite{wu2017comprehensive}, we cited the most competitive results under different experimental settings.} and the proposed GaitSet. Except for GaitSet, the other results were directly taken from their original papers.
All the results were averaged on the 11 gallery views and the identical views were excluded. For example, the accuracy of probe view $36^\circ$ was averaged on 10 gallery views, excluding gallery view $36^\circ$. From Tab. ~\ref{tab:casia-b}, an interesting relationship between views and accuracies can be observed. In addition to $0^\circ$ and $180^\circ$, where low accuracies are expected, the accuracy for $90^\circ$ is a local minimum value that is always worse than the accuracy for $72^\circ$ or $108^\circ$. The possible reason is that gait contains feature information not only those parallel to the walking direction, such as stride, which can be observed most clearly at $90^\circ$ but also feature information vertical to the walking direction, such as the left-right swinging motions of the body or arms, which can be observed most clearly at $0^\circ$ or $180^\circ$. Therefore, both parallel and vertical perspectives lose some portion of the gait information while views such as $36^\circ$ and $144^\circ$ achieve a better balance between these two extremes.

\noindent \textbf{Small-Sample Training~(ST)}~~Our method achieved a high performance even with only 24 subjects in the training set and exceeded the best performance reported~\cite{wu2017comprehensive} by well over $10\%$ on the reported values. There are two main reasons for these results. \textbf{1)} Because our model regards the input as a set, the number of samples (frames) available for training the convolutional network in the main pipeline is dozens of times higher than the number of samples used to train the template- or video-based models. Taking a mini-batch as an example, our model input consists of $30\times 128=3,840$ silhouettes, while under the same batch size other template-based models only obtain $128$ templates. \textbf{2)} Because the sample \textbf{sets} used in the training phase are composed of frames selected randomly from the sequence in the training set, each of which can generate multiple different sample \textbf{sets}; thus, any units related to set feature learning (such as MGP and HPM) can also be trained well.

In addition, it is noteworthy that in ST, all the other compared models were trained and tested only on the NM subset, whereas our model was trained and tested on all the NM, BG, and CL subsets. If we instead focus our model on only one subset, it performs even better, because then, the training and testing environments are the same and exhibit more consistency.

\begin{table}[htb]
\centering
\caption{Averaged rank-1 accuracies on \textbf{OU-MVLP}, excluding identical-view cases. GEINet:~\protect\cite{shiraga2016geinet}. 3in+2diff:~\protect\cite{takemura2017input}}.
\begin{tabularx}{\columnwidth}{|c|c|@{\extracolsep{\fill}}c|c|c|c|}
    \hline
    \multirow{2}{*}{Probe} & \multicolumn{2}{c|}{Gallery All 14 Views} & \multicolumn{3}{c|}{Gallery $0^\circ, 30^\circ, 60^\circ, 90^\circ$} \\
    \cline{2-6}
          & GEINet & Ours & GEINet & 3in+2diff & Ours \\
    \hline
    $0^\circ$     & 11.4  & \textbf{81.3 } & 8.2   & 25.5  & \textbf{79.6 } \\
    \hline
    $15^\circ$    & 29.1  & \textbf{88.6 } & -     & -     & \textbf{87.1 } \\
    \hline
    $30^\circ$    & 41.5  & \textbf{90.2 } & 32.3  & 50.0  & \textbf{87.4 } \\
    \hline
    $45^\circ$    & 45.5  & \textbf{90.7 } & -     & -     & \textbf{89.8 } \\
    \hline
    $60^\circ$    & 39.5  & \textbf{88.6 } & 33.6  & 45.3  & \textbf{86.2 } \\
    \hline
    $75^\circ$    & 41.8  & \textbf{89.1 } & -     & -     & \textbf{88.0 } \\
    \hline
    $90^\circ$    & 38.9  & \textbf{88.3 } & 28.5  & 40.6  & \textbf{84.3 } \\
    \hline
    $180^\circ$   & 14.9  & \textbf{83.1 } & -     & -     & \textbf{81.8 } \\
    \hline
    $195^\circ$   & 33.1  & \textbf{87.7 } & -     & -     & \textbf{84.2 } \\
    \hline
    $210^\circ$   & 43.2  & \textbf{89.4 } & -     & -     & \textbf{87.7 } \\
    \hline
    $225^\circ$   & 45.6  & \textbf{89.7 } & -     & -     & \textbf{87.6 } \\
    \hline
    $240^\circ$   & 39.4  & \textbf{87.8 } & -     & -     & \textbf{86.3 } \\
    \hline
    $255^\circ$   & 40.5  & \textbf{88.3 } & -     & -     & \textbf{86.4 } \\
    \hline
    $270^\circ$   & 36.3  & \textbf{86.9 } & -     & -     & \textbf{85.8 } \\
    \hline
    mean  & 35.8  & \textbf{87.9 } & -     & -     & \textbf{85.9 } \\
    \hline

\end{tabularx}
\label{tab:oumvlp}
\end{table}

\begin{table*}[t]
	\centering
	\caption{Ablation experiments conducted on \textbf{CASIA-B} using setting LT. The results are rank-1 accuracies averaged on all 11 views, excluding identical-view cases. The numbers in brackets indicate the second highest results in each column. Here `att' is the abbreviation of attention.}
	\begin{tabularx}{\textwidth}{@{\extracolsep{\fill}}|c|c|ccccccc|c||c|c|c|}
		\hline
		\multirow{2}{*}{GEI} & \multirow{2}{*}{Set} & \multicolumn{7}{c|}{Set Pooling} & \multirow{2}{*}{MGP} & \multirow{2}{*}{NM} & \multirow{2}{*}{BG} & \multirow{2}{*}{CL} \\
		\cline{3-9}
		&       & Max   & Mean  & Median & Joint sum~\ref{eq:sum} & Joint 1\_1C~\ref{eq:cat} & Pix-att &  Frame att &      &       &       &  \\
		\hline
		$\surd$ &       &       &       &       &       &       &       &      &        & 89.0  & 76.3  & 50.7  \\
		\hline

		& $\surd$ & $\surd$ &       &       &       &       &       &      &       & 95.4  & 88.7  & 69.9 \\
		\hline
		& $\surd$ &       & $\surd$ &       &       &       &       &       &      & 95.0  & 86.3  & 66.3  \\
		\hline
		& $\surd$ &       &       & $\surd$ &       &       &       &       &      & 94.8  & 84.9  & 63.7  \\
		\hline
		& $\surd$ &       &       &       & $\surd$ &       &       &       &      & 94.1  & 84.1  & 64.3  \\
		\hline
		& $\surd$ &       &       &       &       & $\surd$ &       &       &      &94.9  & 86.9 & 66.8  \\
		\hline
		& $\surd$ &       &       &       &       &       & $\surd$ &       &      & 95.6 & 88.9  & 69.6  \\
		\hline
		& $\surd$ &       &       &       &       &       &         & $\surd$ &      & 95.0 & 85.1  & 65.3  \\
		\hline
		& \textbf{$\surd$} & \textbf{$\surd$} &       &       &       &      &       &      & \textbf{$\surd$} & \textbf{96.1 } & \textbf{90.8 } & \textbf{70.3 } \\
		\hline
	\end{tabularx}
	\label{tab:ablation}
\end{table*}

\noindent \textbf{Medium-Sample Training~(MT) \& Large-Sample Training~(LT)}~~
Generally, the performance of deep learning models depends heavily on the scale of the training sets. Thus, we evaluated our GaitSet using two different divisions for the training set and test set, i.e., MT and LT, as recommended in the prior literature.
Tab.~\ref{tab:casia-b} shows that our model attains fairly good results on the NM subset, especially on LT, where the results of all views except $180^\circ$ are over $90\%$. This result shows that the accuracy gains obvious improvement when more training data is available for this subset. 

Our model achieves satisfactory performance on the BG subset.
On the CL dataset, the recognition performances are somewhat less satisfactory, although our model still exceeds the best performance reported so far~\cite{wu2017comprehensive} by over $15\%$. This reduced performance may be explained by the following three reasons: \textbf{1)} a coat can entirely change a person's appearance, e.g., a subject looks larger in a coat than in a T-shirt. \textbf{2)} A coat can hide the motions of both limbs and body. \textbf{3)} In the training set, the ratio of the CL subset is substantially lower than that of the NM subset, demanding stronger discriminative ability on the part of the model.

\subsubsection{OU-MVLP}
Tab.~\ref{tab:oumvlp} compares GaitSet with the two other methods on the OU-MVLP dataset. As some of the previous works did not conduct experiments on all 14 views, we list our results based on two types of gallery sets, i.e. all 14 views and 4 typical views ($0^\circ, 30^\circ, 60^\circ, 90^\circ$). All the results are averaged on the gallery views and identical views are excluded. The results show that our method generalizes well to a dataset with a large scale and wide view variation. 
Moreover, because the representation for each sample only needs to be calculated once, our model can complete the test involving all 133,780 sequences in only 14 minutes with 4 NVIDIA 1080TI GPUs.

It is noteworthy that because some subjects missing several gait sequences have not been removed from the probe set, the maximum rank-1 accuracy cannot reach $100\%$. If we do not count the cases that have no corresponding samples in the gallery, the average rank-1 accuracy of all probe views will rise to $94.1\%$ rather than $87.9\%$.

\begin{figure}[htbp]
\centering
\includegraphics[width=1\linewidth, clip=true, trim=50 250 50 290]{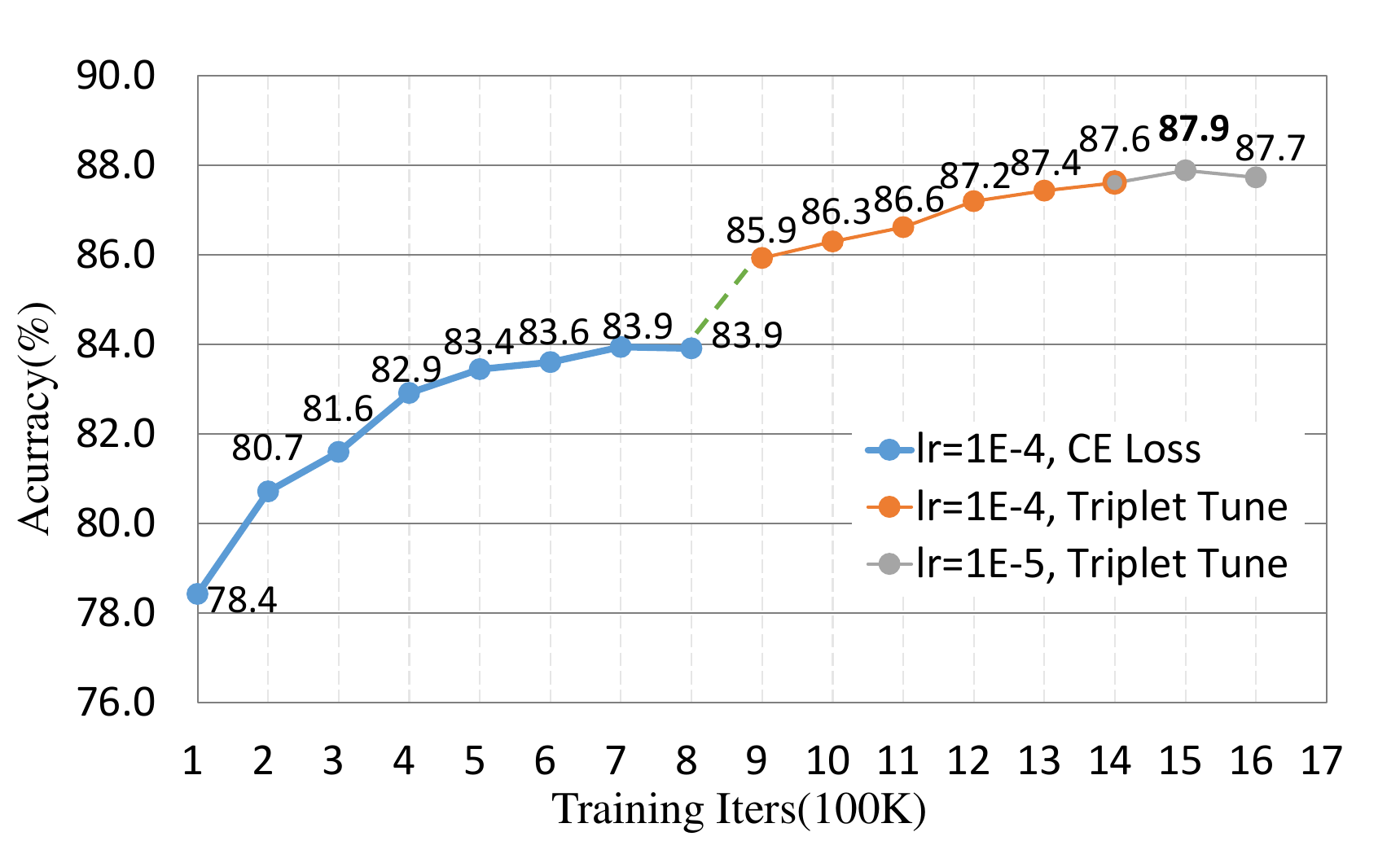}
\caption{The accuracy change process on OU-MVLP.}
\label{fig:ac-ou}
\end{figure}

Fig.~\ref{fig:ac-ou} shows the relationship between training iterations and test accuracy. We can see that after the cross entropy loss reaches its best performance, further tuning with triplet loss can still engender improvement.

\begin{table}[htbp]
\centering
\caption{The impact of different HPM scales and HPM weight independence experiments conducted on \textbf{CASIA-B} using the setting on LT. The results are rank-1 accuracies averaged on all 11 views, excluding identical-view cases.}
    \begin{tabular}{|c|cc|c|c|c|}
    \hline
    \multirow{2}{*}{HPM scales} & \multicolumn{2}{|c|}{HPM weights} & \multirow{2}{*}{NM} & \multirow{2}{*}{BG} & \multirow{2}{*}{CL} \\
    & Shared & Independent &&&\\
    \hline
    1(no HPM) &       & $\surd$  & 91.8  & 82.4  & 59.4  \\    \hline
    2         &       & $\surd$  & 91.8  & 82.9  & 60.1  \\    \hline
    3         &       & $\surd$  & 91.7  & 83.0  & 64.2  \\    \hline
    4         &       & $\surd$  & 93.9  & 86.9  & 64.5  \\    \hline
    \multirow{2}{*}{5}& $\surd$ &      & 91.1  & 82.0  & 60.9  \\ \cline{2-6}
    &         & $\surd$ & 96.1  & 90.8  & 70.3  \\  \hline
    \end{tabular}
  \label{tab:HPM}
\end{table}

\subsection{Ablation Experiments and Model Studies}
\label{sec:ms}

In this section, we report ablation experiments and model studies on CASIA-B, to examine the effectiveness of regarding gait as a set with set pooling, MGP, HPM, and different training strategies with different loss combinations. All the experiments were based on the settings under CASIA-B LT, as shown in Tab.~\ref{tab:casia-b}.

\subsubsection{Ablation experiments}

\textbf{Set VS. GEI}
The first two rows of Tab.~\ref{tab:ablation} show the effectiveness of regarding gait as a set. 
With totally identical networks, the result of using the \textbf{set} exceeds that of using GEI by more than $6\%$ on the NM subset and more than $19\%$ on the CL subset. The only difference is that in the GEI experiment, the gait silhouettes are averaged into a single GEI before being fed into the network.
There might be two main reasons for this improvement:~\textbf{1)}~
our SP extracts the set-level feature from a high-level feature map where the temporal information is well preserved and the spatial information has been sufficiently processed; and \textbf{2)}~as mentioned in Sec.~\ref{sec:mr}, regarding gait as a set enlarges the volume of training data.

\noindent \textbf{The impact of SP}~~In Tab.~\ref{tab:ablation}, the results from the second row to the eighth row show the impact of different SP strategies. SP with pixel-wised attention achieves the highest accuracy on the NM and BG subsets and when $\mathrm{max}(\cdot)$ is used, it obtains the highest accuracy on the CL subsets. Considering the fact that SP with $\mathrm{max}(\cdot)$ also achieves the second best performance on the NM and BG subsets and has the most concise structure; thus, we  choose it as the SP strategy in the final version of GaitSet.

\noindent \textbf{The impact of MGP}~~The second and the last rows of Tab.~\ref{tab:ablation} show that MGP improves all three test subsets. This result is consistent with our experience mentioned in Sec.~\ref{sec:MGP}, i.e., that set-level features extracted from different layers of the main pipeline contain different discriminative information.

\noindent \textbf{The impact of HPM scales and HPM weight independence}~~As shown in Tab.~\ref{tab:HPM}, HPM obtains better performance with more scales. Furthermore, the last two lines of Tab.~\ref{tab:HPM} compare the impact of the weight independence of the fully connected layer in HPM. It can be seen that using independent weights increases the accuracy by more than $7\%$ on each subset. During the experiments, we also noticed that introducing independent weights makes the network converge faster.

\subsubsection{Training strategies}
\label{loss strategies}
Our previous AAAI-19 paper utilized triplet loss to achieve good performance. To further improve the gait recognition accuracy, we combined triplet loss and cross-entropy loss to train the GaitSet model.

\begin{table}[htbp]
\centering
\caption{Different loss functions conducted on \textbf{CASIA-B} using setting LT. The
 results are rank-1 accuracies averaged on all 11 views, excluding identical-view cases. }
    \begin{tabular}{|c|cc|c|c|c|}
    \hline
    \multirow{2}{*}{Loss function} & \multirow{2}{*}{BN} & \multirow{2}{*}{Dropout} & \multirow{2}{*}{NM} & \multirow{2}{*}{BG} & \multirow{2}{*}{CL} \\
    &&&&&\\
    \hline
    \multirow{3}{*}{CEloss} & $\surd$ &         & 90.9  & 84.8  & 46.7  \\
                            &         & $\surd$ & 94.4  & 89.4  & 60.2  \\
                            & $\surd$ & $\surd$ & 95.6  & 90.1  & 64.8  \\
    \hline
    \multirow{2}{*}{Triplet loss} & $\surd$ &      & 95.5  & 87.3  & 69.3 \\
                                &   & $\surd$ & 94.5  & 87.1  & 66.9 \\
                                  & $\surd$ & $\surd$ & 95.3  & 90.3  & 67.3 \\
    \hline
    \multirow{3}{*}{CEloss + Triplet loss} & $\surd$ &         & 95.8  & 90.3  & 69.5   \\
                                           &         & $\surd$ & 94.7  & 88.7  & 66.9 \\
                                           & $\surd$ & $\surd$ & 96.1  & 90.8  & 70.3  \\   
    \hline
    \end{tabular}
  \label{tab:loss function}
\end{table}

Tab.~\ref{tab:loss function} shows the results of the three training strategies, and the impact of batch normalization and dropout. All three training strategies exceed $95\%$ rank-1 accuracy on the NM subset when using batch normalization and dropout. However, only the pretraining model that combines the two losses reaches the highest  $96.1\%$ rank-1 accuracy. 
The first and third lines of Tab.~\ref{tab:loss function} reveal that the dropout layers is essential for a robust training performance of cross-entropy loss in this case. 
We also see that batch normalization improves all the training strategies. 

\subsection{Feature Dimension Reduction}
\label{sec:fdr}
As mentioned in Sec.~\ref{sec:TT}, the testing feature after concatenating all the HPM ouputs has $256\times 31\times 2=15,872$ dimensions in a standard framework, which impairs the testing efficiency. Therefore, we conducted dimensional reduction using two methods. One is to set the feature dimension to a lower level by shrinking the output dimension of the HPM fully connected layers. The other method is to perform the testing task after introducing a new fully connected layer, which achieves a large compression of the original $15,872$ dimensions.

\begin{figure*}[htbp]
\centering
\includegraphics[width=1\linewidth, clip=true, trim=15 80 15 80]{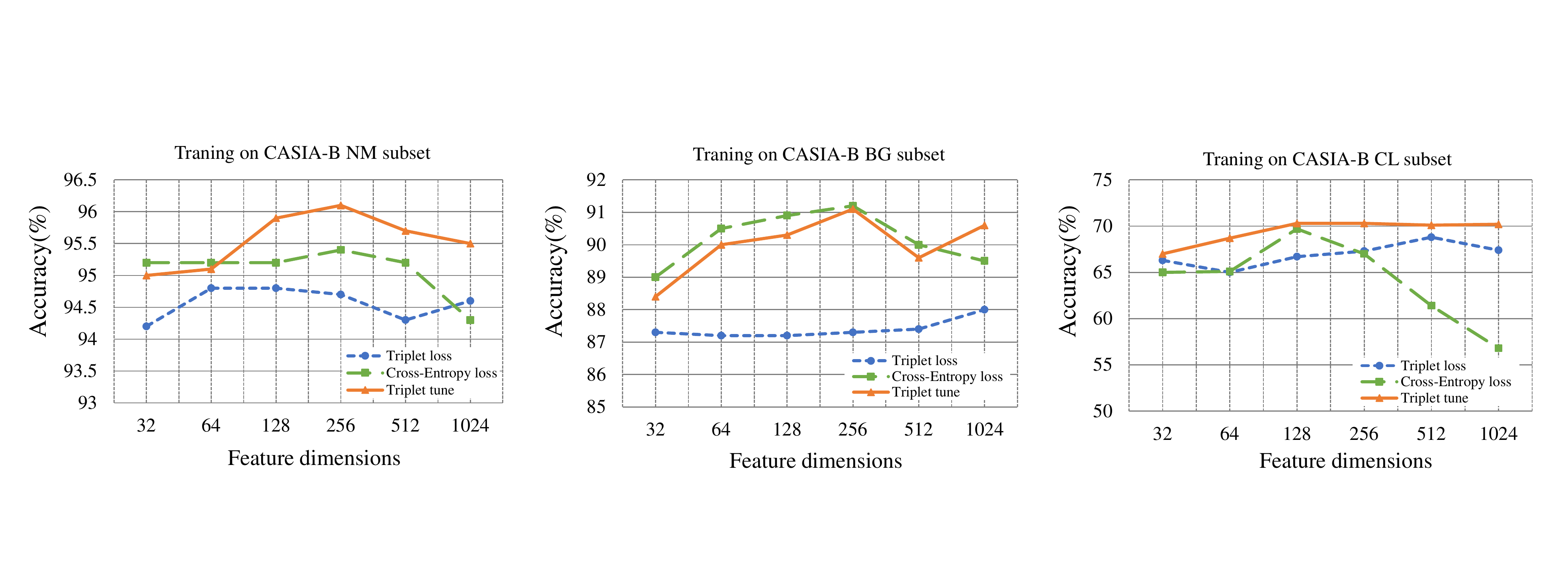}
\caption{Relationships between recognition accuracy and the HPM output dimension. From left to right are the individual results on the CASIA-B NM, BG, and CL subsets. The relationships vary with different training strategies, as shown by the different lines in each graph.}
\label{fig:dim}
\end{figure*}

\noindent \textbf{HPM Output Dimensions.}~~
The HPM output dimensions were set to $32,64,128,256,512$, and $1,024$.
Using these different dimension, we studied the recognition accuracy with approximately $50,000-60,000$ training iterations. As Fig.~\ref{fig:dim} shows, even when the output dimensions are as low as $32$, the performance still reaches $93\%$ on the NM subset with all three loss function strategies. However, there is still a negative impact on the performance if the HPM output dimensions are too low (down to $32$) or too high (up to $1024$). The reasons for this performance degeneration are that 1) the fully connected layers whose output dimensions are too high can easily be overfitting because they contain too many parameters, and 2) an output dimension that is too small would significantly constrain the fully connected layers' learning capacity. In particular, the model trained with CE loss is less robust on the CL subset with high dimensional HMP outputs, while the model with the pretraining strategy has a stable performance on the CL subset. By decreasing the HMP output dimensions, we can compress the final feature dimension from $15,872$ to one quarter of that. While this compression is associated with a subtle performance impairment, the degeneration of recognition performance on the BG and CL subsets cannot be ignored.

\noindent \textbf{Dimension reduction with the new FC}~~
Undoubtedly, we can directly reduce the final feature dimension. After the model has been well trained, a new fully connected layer is applied to the $15,872$ dimension feature to reduce it into a lower dimension. This new layer is tuned for $10,000$ iterations with triplet loss and a learning rate of $1e-4$. We investigate the output dimensions of $128, 256, 512, 1024, 2048$, and $4,096$.

\begin{table}[htbp]
\centering
\caption{The recognition accuracy after dimensional reduction with the new FC.}
    \begin{tabular}{|c|c|c|c|}
    \hline
     \multirow{2}{*}{feature dimension}
      & \multirow{2}{*}{NM} & \multirow{2}{*}{BG} & \multirow{2}{*}{CL} \\
      &&& \\
    \hline
    128         & 91.7  & 83.5  & 62.5  \\
    256         & 94.1  & 87.3  & 66.6  \\
    512         & 94.4  & 88.4  & 68.9  \\
    1024        & 95.0  & 89.3  & 69.1  \\
    2048        & 95.0  & 90.2  & 70.0  \\
    4096        & 94.9  & 90.2  & 70.3  \\
    \hline
    \end{tabular}
  \label{tab:Dimension}
\end{table}

As the Tab.~\ref{tab:Dimension} shows, the final feature dimension can be compressed to $1024$ while maintaining the recognition accuracy at $95\%$ on the NM subset; this is only $6.5\%$ of the original $15,872$ dimensions. Similar to changing the output dimension of HPM, a too-small feature dimension leads to a performance decrease. 
Although it runs counter to the idea of an end-to-end design, introducing this postprocessing effectively compresses the learned feature representation, making the method more practical for real applications.  

\subsection{Practicality}
\label{sec:pra}
Because of the flexibility of the gait set approach, GaitSet may be useful in more complicated practical conditions. In this section, we investigate the practicality of GaitSet through three novel scenarios: \textbf{1)}~How does GaitSet perform when a input set contains only a few silhouettes? \textbf{2)}~Can silhouettes with different views enhance the identification accuracy? \textbf{3)}~Can the model effectively extract deep discriminative representation from a set containing silhouettes shot under different walking conditions? It is worth noting that we did not retrain or tune our model in these experiments; the exact model used in the Sec.~\ref{sec:mr} with the LT setting is used here. Note that here the reported accuracies are averaged on 10 times experiments with different random seeds.

\begin{figure}[htb]
\centering
\includegraphics[width=1\linewidth, clip=true, trim=230 140 230 140]{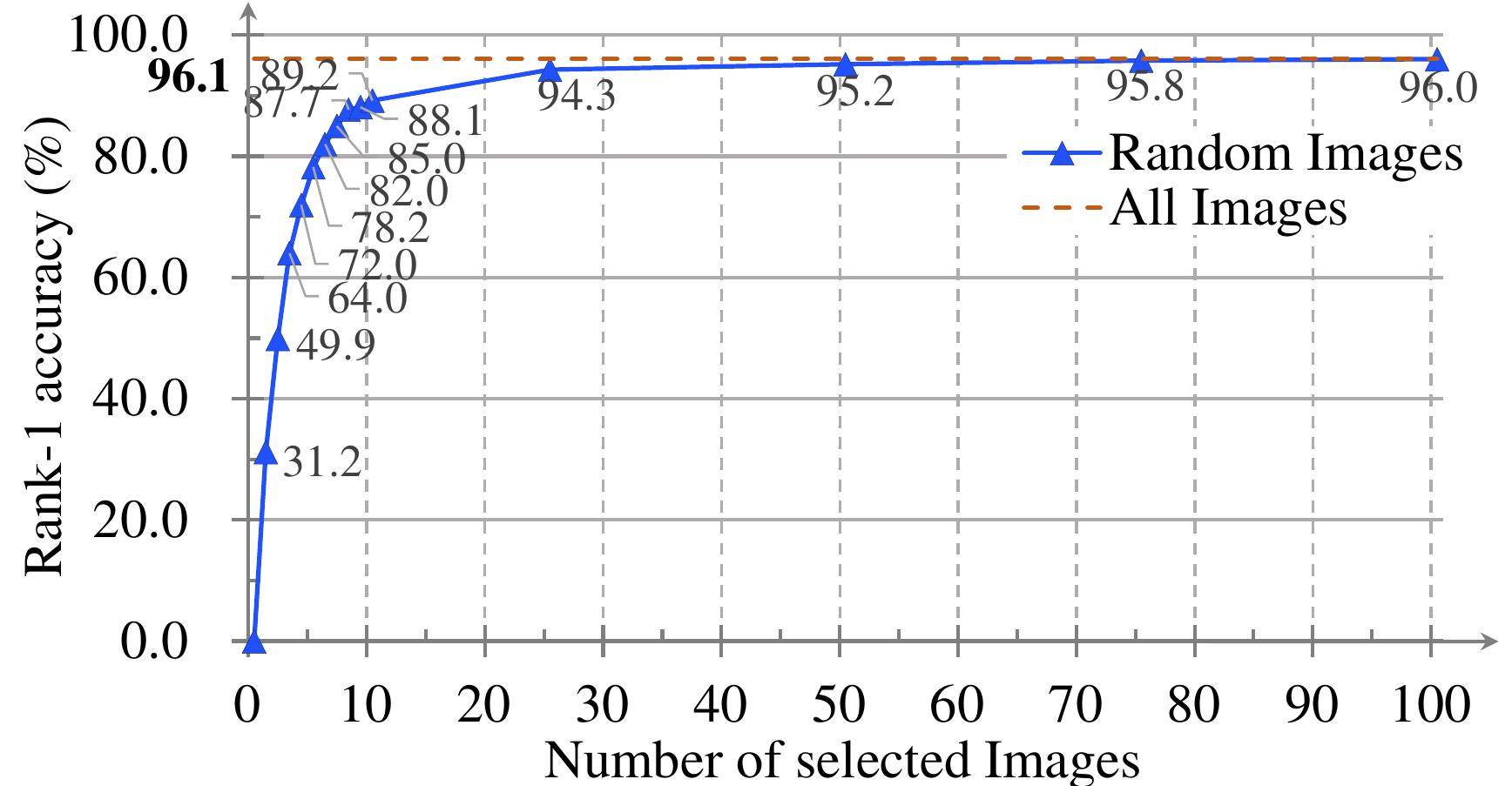}
\caption{Average rank-1 accuracies with constraints of silhouette volume on the \textbf{CASIA-B} dataset using the LT setting. The accuracy values are averaged on all 11 views excluding identical-view cases, and the final reported results are averaged across 10 experimental repetitions.}
\label{fig:frames}
\end{figure}

\noindent \textbf{Limited Silhouettes.}~~In real forensic identification scenarios, cases occur in which no continuous sequence of a subject's gait is available, only some fitful and sporadic silhouettes. We simulate such a circumstance by randomly selecting a certain number of frames from sequences to compose each sample in both the gallery and probe. Fig.~\ref{fig:frames} shows the relationship between the number of silhouettes in each input set and the rank-1 accuracy averaged on all 11 probe views. Our method attains an $82\%$ accuracy using only 7 silhouettes. This result also indicates that our model makes full use of the temporal information contained in a gait set. It can also be observed that \textbf{1)} the accuracy rises monotonically as the number of silhouettes increases, and \textbf{2)} the accuracy is close to the best performance when the samples contain more than 25 silhouettes. This number is consistent with the number of frames that one gait period contains.

\noindent \textbf{Multiple Views.}~~
Here we study a scenario where one person's gait is collected from different views. 
We simulate these scenarios by constructing each silhouette sample selected from two sequences that have the same walking condition but different views. To alleviate the effects of the number of silhouettes, an experiment is performed under the case where the maximum silhouette number is 10. Unlike the previous contrast experiments of a single view where an input set consists of 10 silhouettes from one sequence, more concretely, in this experiment, an input set is made up of 5 silhouettes from each of two sequences in the two-view experiment. Note that in this experiment, only probe samples are composed by the aforementioned method, whereas the sample in the gallery is composed of all silhouettes from one sequence.

\begin{table}[htbp]
\centering
\caption{Multiview experiments conducted on \textbf{CASIA-B} using the LT setting. Cases with the probe contains the views in the gallery are excluded.}
    \begin{tabular}{|c|c|c|c|c|c||c|}
    \hline
    \multirow{2}{*}{\shortstack{View \\ difference}} &
    \multirow{2}{*}{\shortstack{$18^\circ$/\\$162^\circ$}} &
    \multirow{2}{*}{\shortstack{$36^\circ$/\\$144^\circ$}} &
    \multirow{2}{*}{\shortstack{$54^\circ$/\\$126^\circ$}} &
    \multirow{2}{*}{\shortstack{$72^\circ$/\\$108^\circ$}} &
    \multirow{2}{*}{$90^\circ$} &
    \multirow{2}{*}{\shortstack{Single \\ view}} \\
    &&&&&&\\
    \hline
    All silhouettes & 98.9  & 99.6  & 97.6  & 96.2  & 99.3  & 96.1  \\
    \hline
    10 silhouettes  & 94.8  & 97.2  & 95.9  & 91.7  & 97.25  & 89.5  \\
    \hline
    \end{tabular}%
  \label{tab:mv}%
\end{table}%

Because there are too many view pairs to display them all, we summarize the results by averaging the accuracies of each possible view difference, as indicated in Tab.~\ref{tab:mv}. For example, the result of a $90^\circ$ difference was averaged by the accuracies of $6$ view pairs~($0^\circ\& 90^\circ, 18^\circ\& 108^\circ, ..., 90^\circ\& 180^\circ$). Furthermore, the 9 view differences were folded at $90^\circ$ and those larger than $90^\circ$ were averaged with the corresponding view differences of less than $90^\circ$. For example, the results of the $18^\circ$ view difference were averaged with those of the $162^\circ$ view difference.Our model effectively aggregates information from different views and boosts the final performance.
This result can be explained by the pattern between views and accuracies discussed in Sec.~\ref{sec:mr}. Including multiple views in the input set allows the model to gather both parallel and vertical information, resulting in performance improvements.

\textbf{Multiple Walking Conditions.}~~
In real life, it is highly likely that gait sequences of the same person
 occur under different walking conditions.
 We simulate such different conditions by forming an input set using silhouettes from two sequences with the same view but different walking conditions and conduct experiments under the constraint of different numbers of silhouettes. Note that in this experiment, only the probe samples are composed by the method discussed above. All the samples in the gallery are constructed using all the silhouettes from one sequence. Moreover, the probe-gallery division of this experiment is different.
 For each subject, the sequences NM \#02, BG \#02, and CL \#02 are kept in the gallery, while the sequences NM \#01, BG \#01, and CL \#01 are used as probes.

\begin{table}[htbp]
\centering
\caption{Multiple walking condition experiments conducted on \textbf{CASIA-B} using the LT setting. The results are rank-1 accuracies averaged on all 11 views, excluding identical-view cases. The numbers in brackets indicate the constraints of the silhouette number in each input set.}
    \begin{tabularx}{\columnwidth}{@{\extracolsep{\fill}}|c|c|c|c|c|c|}
    \hline
    NM(10) & 91.1  & NM(10)+BG(10) & 93.0  & NM(20) & 94.2  \\
    \hline
    BG(10) & 85.8  & NM(10)+CL(10) & 91.6  & BG(20) & 89.4  \\
    \hline
    CL(10) & 85.1  & BG(10)+CL(10) & 89.7  & CL(20) & 88.9  \\
    \hline
    \end{tabularx}%
  \label{tab:mw}%
\end{table}%

From Tab.~\ref{tab:mw} we can see that 1) the accuracies are still boosted as the number of silhouettes increases, and 2) when the number of the silhouettes is fixed, the results reveal the relationships between different walking conditions. Containing large yet complementary noises and information, the combination of silhouettes from BG and CL helps the model improve the accuracy. In contrast, silhouettes of NM contain little noise. Consequently, substituting silhouettes of the other two conditions for some of them does not provide extra useful information but only introduces noise, leading to a degraded performance.

\ifx\allinone\undefined
\newpage
\bibliographystyle{aaai}
\bibliography{../ref}
\end{document}
\fi

\ifx\allinone\undefined
\input{../preamble}
\begin{document}
\fi

\section{Conclusion}
\label{sec:con}
In this paper, we presented a novel perspective that regards gait as a deep set, called a GaitSet. The proposed GaitSet approach extracts both spatial and temporal information more effectively and efficiently than do the existing methods, which regard gait as either a template or a sequence.
Unlike other existing gait recognition approaches, the GaitSet approach also provides an innovative way to aggregate valuable spatiotemporal information from different sequences to enhance the accuracy of cross-view gait recognition. 
Experiments on two benchmark gait datasets indicate that GaitSet achieves the highest recognition accuracy compared with other state-of-the-art algorithms, and the results reveal that GaitSet exhibits a wide range of flexibility and robustness when applied to various complex environments, showing a great potential for practical applications. In addition, since the set assumption could fit various other biometric identification tasks including person re-identification and video-based face recognition, the structure of GaitSet can be applied to these tasks with few minor changes in the future.

\ifx\allinone\undefined
\end{document}
\fi

\section*{Acknowledgement}
The authors would like to thank the associate editor and
anonymous reviewers for their valuable comments,
which greatly improved the quality of this paper.

\bibliographystyle{IEEEtran}
\bibliography{ref}

\begin{IEEEbiography}[{\includegraphics[height=1.2in,clip,keepaspectratio,trim=0 0 0 0]{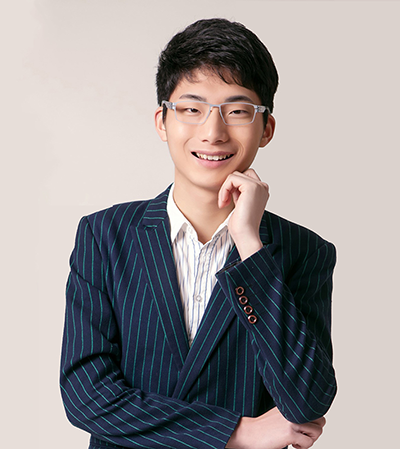}}]{Hanqing Chao} received the B.S. degree from the Nanjing University of Aeronautics and Astronautics, Nanjing, China, in 2016. He received the Master's degree in Computer Science at Fudan University, Shanghai, China, in 2019. His research interests include machine learning, computer vision and gait recognition.
\end{IEEEbiography}

\begin{IEEEbiography}[{\includegraphics[height=1.2in,clip,keepaspectratio,trim=0 0 0 0]{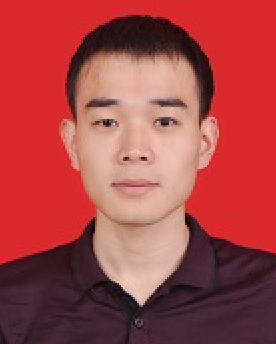}}]{Kun Wang} received the B.S. degree from the Xi'an Jiaotong University, Xi'an, China, in 2017. He received the Master's degree in Computer Science at Fudan University, Shanghai, China in 2020. His research interests include machine learning, computer vision and gait recognition.
\end{IEEEbiography}

\begin{IEEEbiography}[{\includegraphics[height=1.2in,clip,keepaspectratio,trim=0 0 0 100]{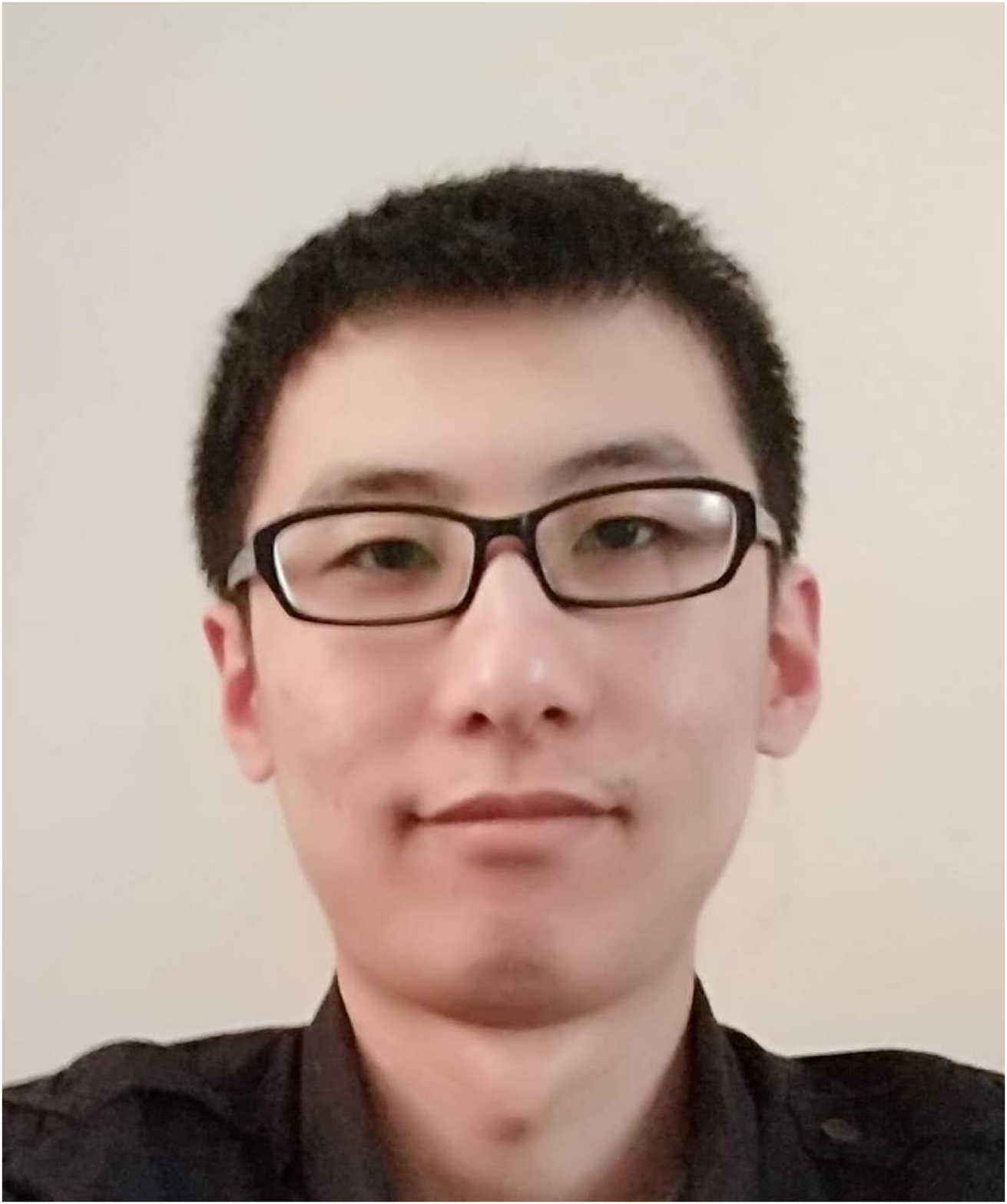}}]{Yiwei He} received the B.S. degree from the Nanjing Audit University, Nanjing, China, in 2015. He received the Master's degree in Computer Science at Fudan University, Shanghai, China, in 2018. His research interests include machine learning, computer vision and gait recognition.
\end{IEEEbiography}

\begin{IEEEbiography}[{\includegraphics[height=1.2in,clip,keepaspectratio,trim=0 0 0 0]{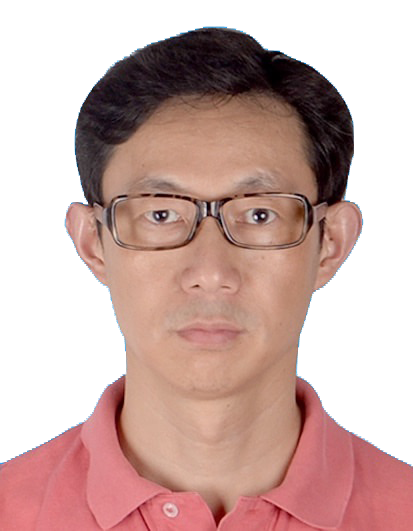}}]{Junping Zhang}  (M'05) received the B.S. degree in automation from Xiangtan University, Xiangtan, China, in 1992. He received the M.S. degree in control theory and control engineering from Hunan University, Changsha, China, in 2000. He received his Ph.D. degree in intelligent system and pattern recognition from the Institute of Automation, Chinese Academy of Sciences, in 2003. He is a professor at School of Computer Science, Fudan University since 2011. His research interests include machine learning, image processing, biometric authentication, and intelligent transportation systems. He has widely published in highly ranked international journals such as IEEE TPAMI and IEEE TNNLS, and leading international conferences such as ICML and ECCV. He has been an Associate Editor of IEEE Intelligent Systems Magazine since 2009.
\end{IEEEbiography}

\begin{IEEEbiography}[{\includegraphics[height=1.2in,clip,keepaspectratio,trim=100 0 0 0]{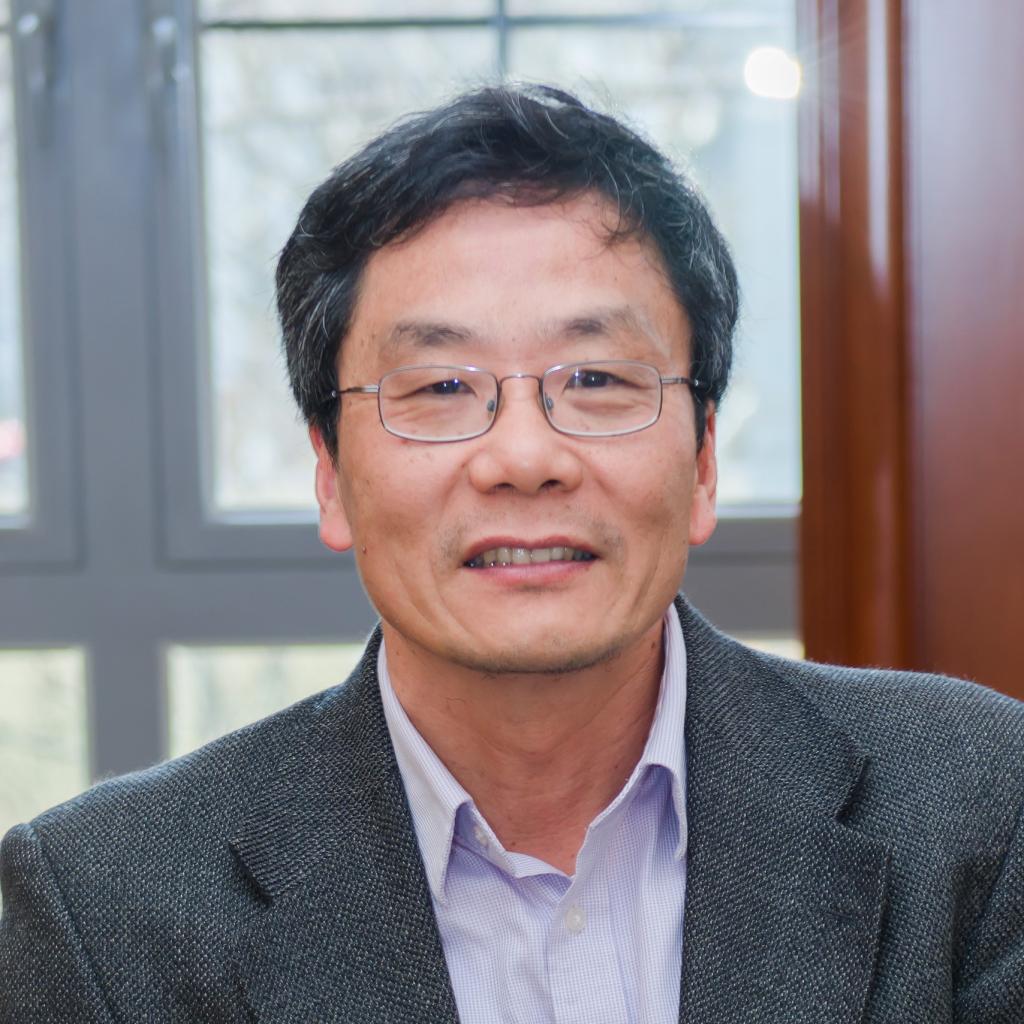}}]{Jianfeng Feng} received all his academic degrees from Peking University in mathematics, Peking, China, in 1993. He is the chair professor of Shanghai National Centre for Mathematic Sciences, and the Dean of Brain-inspired AI Institute and the head of Data Science School in Fudan University since 2008. He has been developing new mathematical, statistical and computational theories and methods to meet the challenges raised in neuroscience, mental health and brain-inspired AI researches.  He was awarded the Royal Society Wolfson Research Merit Award in 2011, as a scientist ‘being of great achievements or potentials’. He was invited to deliver 2019 Paykel Lecture at the Cambridge University. 
\end{IEEEbiography}

\end{document}